\documentclass[10pt,twocolumn,letterpaper]{article}

\usepackage{arxiv}      

\usepackage{booktabs}
\usepackage{multirow}
\usepackage{diagbox}
\usepackage{pgfplots}
\usepackage{tcolorbox}

\definecolor{iccvblue}{rgb}{0.21,0.49,0.74}
\usepackage[pagebackref,breaklinks,colorlinks,allcolors=iccvblue]{hyperref}
\usepackage[capitalize]{cleveref}

\newcommand\mypara[1]{\vspace{1.0mm}\noindent\textbf{#1}}

\def\paperID{1897} 
\def\confName{ICCV}
\def\confYear{2025}

\title{Cockatiel: Ensembling Synthetic and Human Preferenced Training for Detailed Video Caption}

\author{
Luozheng Qin $^{1}$ \quad
Zhiyu Tan $^{1,2}$ \quad
Mengping Yang \quad 
Xiaomeng Yang $^{1}$ \quad 
Hao Li $^{1,2}$\footnotemark[2] \\ 
{
$^{1}$ {Shanghai Academy of Artificial Intelligence for Science} \quad 
$^{2}$ {Fudan University}
}  \\
{
qinluozheng@sais.com.cn
}\\
Project: \href{https://sais-fuxi.github.io/projects/cockatiel}{https://sais-fuxi.github.io/projects/cockatiel}
} 
\begin{document}

\twocolumn[{%
    \renewcommand\twocolumn[1][]{#1}%
    \maketitle
    \begin{center}
        \centering
        \includegraphics[width=0.93\linewidth]{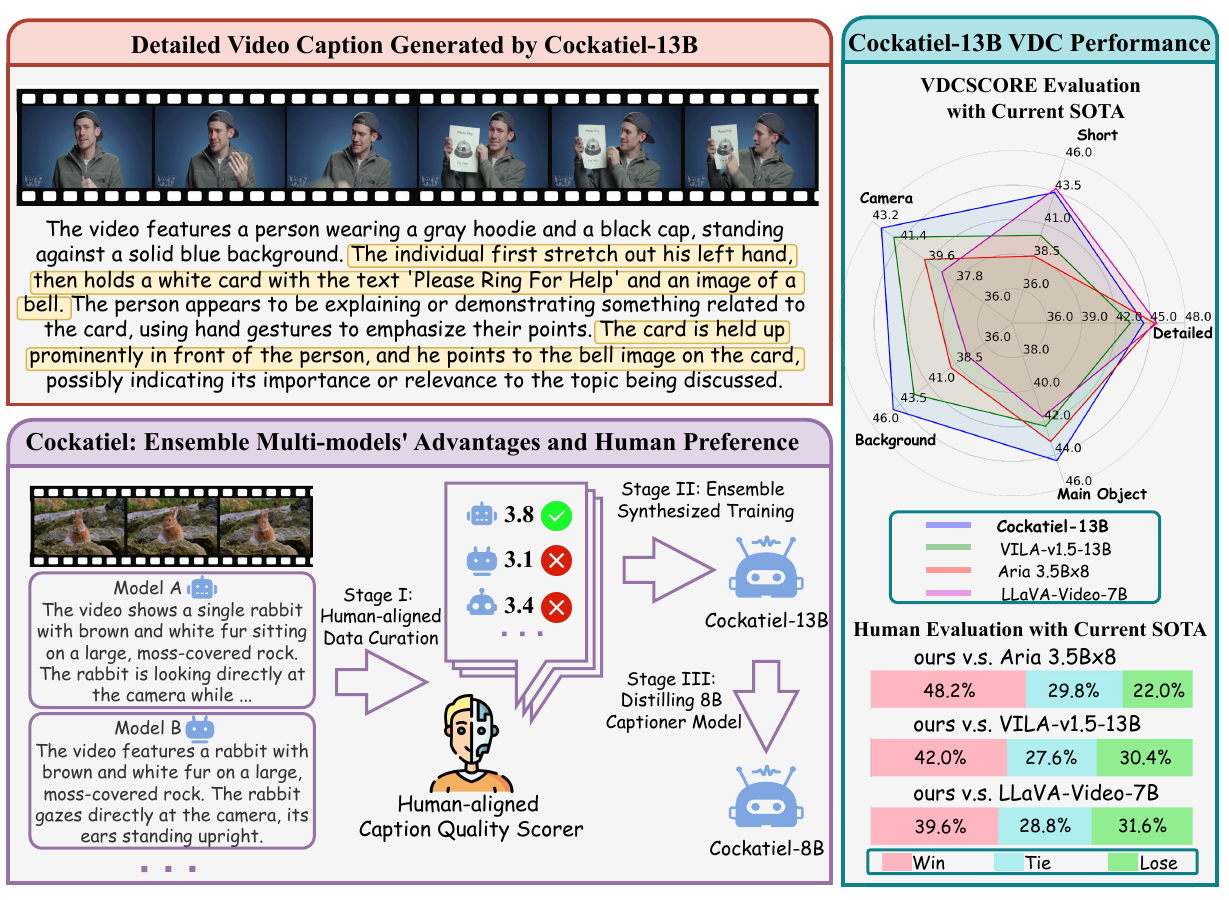}
        \vspace{-7pt}
        \captionof{figure}{ 
            \textbf{Cockatiel, a three-stage training pipeline that ensembles synthetic and human-aligned training.}
            Cockatiel-13B are capable to generate detailed captions that consistently aligns with every visual element in the input video (\textit{top left}).
            Furthermore, Cockatiel-13B achieves new state-of-the-art and considerable dimension-balanced performance on VDCSCORE while consistently voted as the most human-aligned models compared to baselines (\textit{right}). 
            The key contributor to these capabilities is the ensembling synthetic and human preferenced training, which infuses Cockatiel-13B with diverse strengths of leading VDC models and human preferences (\textit{bottom left}).
        }
        \label{fig:teaser}
    \end{center}  
}]

\maketitle
\begin{abstract}

Video Detailed Captioning (VDC) is a crucial task for vision-language bridging, enabling fine-grained descriptions of complex video content.
In this paper, we first comprehensively benchmark current state-of-the-art approaches and systematically identified two critical limitations: biased capability towards specific captioning aspect and misalignment with human preferences.
To address these deficiencies, we propose \textbf{Cockatiel}, a novel three-stage training pipeline that ensembles synthetic and human-aligned training for improving VDC performance.

In the first stage, we derive a scorer from a meticulously annotated dataset to select synthetic captions high-performing on certain fine-grained video-caption alignment and human-preferred while disregarding others.
Then, we train Cockatiel-13B, using this curated dataset to infuse it with assembled model strengths and human preferences.
Finally, we further distill Cockatiel-8B from Cockatiel-13B for the ease of usage.
Extensive quantitative and qualitative experiments reflect the effectiveness of our method, as we not only set new state-of-the-art performance on VDCSCORE in a dimension-balanced way but also surpass leading alternatives on human preference by a large margin as depicted by the human evaluation results.
\end{abstract}
\section{Introduction}
Video Detailed Captioning~(VDC) aims to generate comprehensive and detailed captions that capture the visual semantics and temporal dynamics of input videos.
This capacity serves as a fundamental ability for the powerful Multimodal Large Language Models~(MLLMs)~\citep{li2023blip2, zhu2023minigpt4, yue2024mmmu} and provides nuanced textual guidance for the emerging video diffusion models~\citep{ma2024latte, zheng2024opensora, yang2024cogvideox}, facilitating various video-related understanding and generation tasks, such as video question answering~\citep{xu2017videoqa, yu2018joint}, video retrieval~\citep{qin2024bvrcc, jiang2022tencent_t2v_retrieval}, text-to-video generation~\citep{gupta2024walt, li2024hunyuan_dit}, text-guided video editing~\citep{ceylan2023pix2video, he2024id-animator}, \emph{etc.}
In general, video detailed captioning is pivotal for the performance and applicability of video-based AI systems.


Unfortunately, through our comprehensive benchmark of current VDC models on VDCSCORE~\citep{chai2024auroracap}, we observe that existing VDC models are hindered by imbalanced fine-grained video-caption alignment and misalignment with human preference.
Specifically, a single VDC model usually prefers specific aspects of VDC rather than providing comprehensive captions, as it is challenging to find a model that performs collectively competitive across the evaluated dimensions of VDCSCORE.
Besides, since existing VDC models are mainly trained on purely synthetic detailed image/video captions~\citep{chen2024panda-70m, li2024llava-ov, kim2024generalizing_vqa}, they are unable to align well with human preferences.
Nevertheless, to mitigate these challenges, we can ensemble data generated by various VDC models to leverage their respective strengths and incorporate human preferences into the ensemble for improving the alignment with them.
 
However, to ensure the effective integration of diverse model strengths and human preferences, it is crucial to accurately select and involve the most advantageous one for training from captions produced by various models.
Besides, we have to prioritize captions preferred by humans while disregarding those are disapproved, thereby improving the alignment with human preference.
Consequently, a robust, human-aligned selection policy is critical for our method, as it determines that, at the instance level, which model's generated captions should be involved in training based on dimension-specific performance and human preference.
To achieve this, we employ an innovative selection policy powered by a human-aligned caption quality scorer, which is essentially a MLLM fine-tuned on a meticulously annotated dataset of structured human preference score on VDC.
Our selection policy not only selects the highest-scored caption among its candidates for each video, but also applies a threshold setting to filter out cases where all candidates are low in quality.
Then, we  use this curated dataset to fine-tune VDC models, thereby obtaining a competitive, dimension-balanced, and human-preferred VDC models.
For the ease of user usage and deployment, We further distill Cockatiel-8B by utilizing Cockatiel-13B as an additional base model.
Cockatiel significantly alleviates the aforementioned challenges, achieving new state-of-the-art and considerable dimension-balanced performance on VDCSCORE, as depicted in \cref{fig:teaser}.
Notably, to the best of our knowledge, this is the first work that focuses on the imbalanced detailed video-caption alignment and human preference misalignment in VDC, filling a critical gap in the field.
To summarize, our primary contributions are:
\begin{itemize}[itemsep=5pt,topsep=5pt,leftmargin=15pt]

\item By comprehensively benchmarking existing alternatives, we identify two critical challenges of VDC: the imbalanced detailed video-caption alignment and the misalignment with human preference.

\item We propose \textbf{Cockatiel}, a novel framework that ensembles synthetic and human preferenced training on VDC models to address the above challenges.

\item We carefully annotate a dataset with structured human preference scores on detailed video captions and train a human-aligned scorer on it, facilitating high-quality data selection.

\item Extensive experiments validate that our method sets a new state-of-the-art performance on VDCSCORE while achieving better alignment with human preference.

\end{itemize}
\section{Benchmarking Existing Alternatives}
\begin{table*}[t]
\centering
\small
\begin{tabular}{l|cccccc}
\toprule
\multicolumn{1}{c|}{Model}                                                        & \begin{tabular}[c]{@{}c@{}} Camera\\ (Acc / Score)\end{tabular} & \begin{tabular}[c]{@{}c@{}}  Short \\ (Acc / Score)\end{tabular} & \begin{tabular}[c]{@{}c@{}}  Background\\ (Acc / Score)\end{tabular} & \begin{tabular}[c]{@{}c@{}} Main Object\\ (Acc / Score)\end{tabular} & \begin{tabular}[c]{@{}c@{}} Detailed\\ (Acc / Score)\end{tabular} & \begin{tabular}[c]{@{}c@{}} Average\\ (Acc / Score)\end{tabular} \\ 
\midrule
ShareGPT4Video-8B~\citep{chen2024sharegpt4video}  & 29.26/1.54 & 32.60/1.70 & 30.59/1.60 & 30.67/1.61 & 31.67/1.66 & 30.96/1.62 \\
Vriptor~\citep{yang2024vript}            & 37.64/1.96 & 38.35/2.00 & 37.11/1.94 & 37.02/1.93 & 38.49/2.00 & 37.72/1.97 \\
AuroraCap-7B~\citep{chai2024auroracap}       & 33.88/1.77 & 37.98/1.97 & 33.96/1.77 & 35.95/1.87 & 41.52/2.15 & 36.66/1.91 \\ 
\midrule
VideoChatGPT~\citep{maaz2023video-chatgpt}       & 33.19/1.74 & 35.68/1.87 & 33.77/1.76 & 33.54/1.75 & 34.70/1.81 & 34.18/1.79 \\
Video-LLaVA~\citep{lin2023video-llava}        & 32.80/1.72 & 36.23/1.89 & 32.86/1.70 & 33.08/1.73 & 36.47/1.89 & 34.29/1.79 \\
LLaMA-Vid~\citep{zhang2024llava-vid}          & 36.12/1.88 & 38.04/1.98 & 34.27/1.79 & 35.42/1.84 & 35.89/1.87 & 35.95/1.87 \\
PLLaVA-7B~\citep{xu2024pllava}          & 34.45/1.79 & 34.19/1.78 & 34.10/1.78 & 34.40/1.78 & 36.97/1.92 & 34.82/1.81 \\
PLLaVA-13B~\citep{xu2024pllava}         & 37.37/1.93 & 38.10/1.97 & 37.20/1.92 & 36.40/1.89 & 39.68/2.05 & 37.75/1.95 \\
Idefics2-8B~\citep{laurenccon2025idefics2}        & 25.36/1.36 & 34.38/1.78 & 31.80/1.66 & 31.73/1.65 & 31.49/1.63 & 30.95/1.62 \\
VILA-v1.5-8B~\citep{lin2024vila}       & 39.74/2.06 & 39.29/2.04 & 39.84/2.06 & 40.85/2.11 & 42.80/2.21 & 40.50/2.10 \\
VILA-v1.5-13B~\citep{lin2024vila}      & 41.81/2.16 & 40.18/2.08 & 42.27/2.18 & 42.27/2.18 & 43.26/2.23 & 41.96/2.17 \\
NVILA-15B~\citep{liu2024nvila}          & 31.69/1.64 & 40.83/2.11 & 32.80/1.71 & 33.45/1.74 & 42.04/2.17 & 36.16/1.87 \\
VideoChat2-7B~\citep{li2024videochat2}      & 31.94/1.68 & 40.23/2.08 & 34.88/1.82 & 34.93/1.82 & 40.48/2.11 & 36.49/1.90 \\ 
InternVL-v2.5-8B~\citep{chen2024internvl2_5}   & 34.36/1.81 & 42.25/2.19 & 37.58/1.95 & 38.23/1.99 & 42.99/2.22 & 39.08/2.03 \\
LLaMA3.2-Vision-11B~\citep{meta2024llama3_2} & 33.62/1.74 & 35.66/1.83 & 35.44/1.83 & 34.80/1.80 & 36.77/1.90 & 35.26/1.82 \\
InternVideo-v2.5~\citep{wang2025internvideo2}   & 31.37/1.65 & 39.60/2.05 & 32.57/1.70 & 35.19/1.84 & 38.47/1.99 & 35.44/1.85 \\
mPLUG-Owl-Video~\citep{ye2024mplug-owl2}    & 38.17/1.99 & 40.18/2.09 & 37.15/1.95 & 38.49/2.00 & 40.25/2.10 & 38.85/2.03 \\
QwenVL-v2-8B~\citep{wang2024qwen2-vl}       & 35.40/1.85 & 40.80/2.11 & 38.87/2.01 & 40.59/2.10 & 43.56/2.26 & 39.84/2.07 \\
Aria-3.5Bx8~\citep{li2024aria}        & 39.84/2.07 & 38.61/2.01 & 38.97/2.02 & 43.21/2.23 & \underline{45.36/2.33} & 41.20/2.13 \\   
LLaVA-OneVision-7B~\citep{li2024llava-ov} & 37.57/1.96 & 41.65/2.15 & 34.31/1.79  & 38.81/2.02 & 41.81/2.16 & 38.83/2.02 \\
LLaVA-Video-7B~\citep{zhang2024llava-vid}     & 38.73/2.02 & \underline{43.75/2.26} & 37.50/1.95  & 41.71/2.16 & \textbf{45.56/2.35} & 41.45/2.15 \\
\midrule
Cockatiel-8B (Distilled) & \underline{42.25/2.19} & \textbf{44.01/2.27} & \underline{43.89/2.26} & \underline{43.85/2.26} & 44.00/2.27 & \underline{43.60/2.25} \\
Cockatiel-13B            & \textbf{42.62/2.21} & 43.45/2.25 & \textbf{44.13/2.28} & \textbf{44.37/2.29} & 44.42/2.29 & \textbf{43.80/2.26} \\
\bottomrule
\end{tabular}
\vspace{-2pt}
\caption{
\textbf{Quantitative Comparison of existing VDC models on VDCSCORE Benchmark}. 
We incorporate most up-to-date specialized VDC models as well as general MLLMs, since both regard video detailed captioning as their key capability.
For each video, we extract 32 frames for most models, with exceptions for AuroraCap-7B (16), NVILA-15B (16), LLaMA3.2 (16), and Video-LLaVA (8).
This difference ensures every model can fit and perform inference on a single GPU.
The best and second-best results are emphasized using \textbf{bold} and \underline{underline}. 
}
\label{tab:vdc_benchmark_results}
\end{table*}
\subsection{Preliminary}

\mypara{VDCSCORE.}
VDCSCORE~\citep{chai2024auroracap} evaluates the similarity between the predicted and ground-truth detailed captions.
The evaluation process involves three steps.
First, it decomposes the ground-truth caption into a set of concise question-answer pairs using a LLM, then generates corresponding responses from the predicted caption. 
Finally, the LLM is used to assess the accuracy of each response to provide an overall score.
VDCSCORE evaluates how the generated captions describe the objects, backgrounds, camera movements involved in the video, how VDC models generate short and detailed captions, dubbed as ``Main Object'', ``Background'', ``Camera'', ``Short'' and ``Detailed'' in \cref{tab:vdc_benchmark_results}.

\mypara{Current popular video caption models.}
We involve two types of models for evaluation, specialized video detailed captioning models and general video MLLMs, both of which treat video detailed captioning as their fundamental ability.
Specifically, for the specialized VDC models, we employ Vriptor~\citep{yang2024vript}, ShareGPT4Video~\cite{chen2024sharegpt4video}, and AuroraCap~\citep{chai2024auroracap}.
As for the general video MLLMs, we include both pioneer early works in the field, such as VideoChatGPT~\citep{maaz2023video-chatgpt}, Video-LLaVA~\citep{lin2023video-llava, liu2023llava}, LLaMA-Vid~\citep{li2024llama-vid}, PLLaVA~\citep{xu2024pllava}, and latest versions of popular MLLM series, such as Idefics2~\citep{laurenccon2025idefics2}, VILA v1.5~\cite{lin2024vila}, NVILA~\cite{liu2024nvila}, VideoChat2~\citep{li2024videochat2}, InternVL2.5~\citep{chen2024internvl, chen2024internvl2_5}, LLaMA3.2~\citep{dubey2024llama3, meta2024llama3_2}, InternVideo-v2.5~\citep{wang2025internvideo2}, mPLUG-Owl-Video~\citep{xu2023youku-mplug, ye2024mplug-owl2}, QwenVL-v2~\citep{wang2024qwen2-vl}, Aria~\citep{li2024aria},  LLaVA-OneVision~\citep{liu2024llava_1_5, li2024llava-ov}, LLaVA-Video~\citep{zhang2024llava-vid}.
Particularly, we focus on models with fewer than 34B parameters, and uniformly extract 32 frames for input videos. 
By applying such settings, we eliminate the need for model parallelism while preserving more temporal information during inference.

\subsection{Benchmark Results and Analysis}

\mypara{One single caption model usually prefers specific aspects than providing comprehensive results.}
To generate detailed and comprehensive captions, the model must effectively capture all visual elements in the video, including the main objects, camera movements, and backgrounds. 
This requires the model-generated captions to align with videos with good accuracy and coverage in every aspect.
Nevertheless, as evidenced by the evaluation results in \cref{tab:vdc_benchmark_results}, no model demonstrates uniformly competitive performance across all five evaluated dimensions.
Instead, the evaluated models tend to excel in specific aspects while underperforming in others, resulting in captions with uneven fine-grained caption-video alignment across the dimensions.
For instance, LLaVA-OneVision and LLaVA-Video show strong performance in VDCSCORE on ``Short'' and ``Detailed'' but perform inadequately in ``Camera'', while Aria exhibits significant advantages in ``Detailed'' and ``Main Object'' but performs poorly in other aspects.
This imbalanced fine-grained video-caption alignment is especially problematic, as all five evaluated aspects are equally critical and fundamental for the effective employment of VDC models, highlighting a pressing need for improvement.

\mypara{Human alignment is not involved in existing approaches.}
Thorough manual observation on the generated captions, we also noticed they often fails to align with human preference.
Such misalignment can be attributed to the lack of publicly available human-annotated or human-aligned VDC datasets.
As a consequence, VDC models are mainly trained on purely synthesized detailed image/video captions which lacks the guidance of human preference.
While training on data generated by commercial models (such as GPT4-o~\citep{openai2024gpt4o}) might be a solution, it suffers from the task gap between general visual dialogue and VDC, as well as the cost and efficiency of API calling.

\mypara{Ensembling captions from multiple VDC models with human guidance could improve VDC performance.}
As a result of these challenges, there is an urgent need of a VDC model capable of providing comprehensive, detailed, and human-aligned captions that capture every visual element in the videos. 
To fulfill this need, we can ensemble the most dimension-specific advantageous and human-preferred one from candidate captions.
Following this approach in a dimension-to-dimension way, we will obtain a synthetic training dataset that fully leverages the strengths of each ensembled base model while maintaining considerable alignment with human preferences.
Such a dataset can serve as excellent training material for VDC models, featuring high in both dimension-specific video-caption alignment and human preference alignment.
In view of this circumstance, we propose Cockatiel, a three-stage training pipeline that implement the ensemble synthetic and human preferenced training for VDC models.

\begin{figure*}[t]
    \centering
    \includegraphics[width=.8\linewidth]{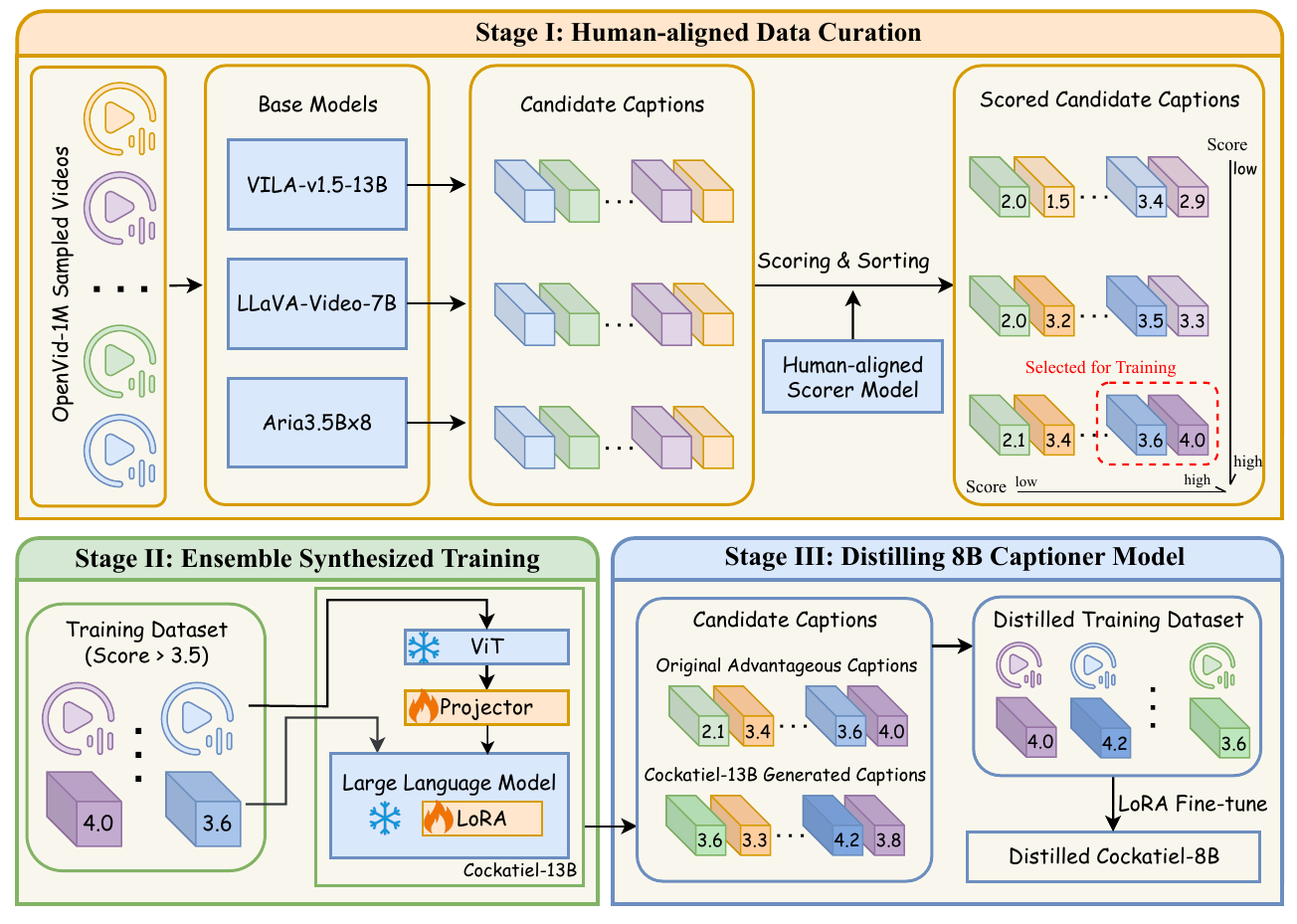}
    \vspace{-7pt}
    \caption{
    \textbf{Overall pipeline of our proposed Cockatiel}. 
    Our training pipeline successfully ensemble both the advantages of base models and human preferences, yielding our Cockatiel captioner series.
    Through the ensembling synthetic and human preferenced training, Cockatiel-13B achieves significant VDC performance while being preferred by humans. 
    }
    \label{fig:pipeline}
\end{figure*}

\section{The Proposed Cockatiel}
\label{sec:method}

\subsection{Human-alignment Caption Scorer \& Selector}
\label{subsec:scorer}

We meticulously annotate a dataset of structured human preference score on video detailed captions and fine-tune a MLLM on it.
In this way, we obtain a human-aligned scorer that can determine the caption training value for the selection policy.
We will introduce details about the data annotation and scorer training in \cref{subsubsec:data_annotation} and \cref{subsubsec:scorer_implementation}. 


\subsubsection{Scoring Data Annotation}
\label{subsubsec:data_annotation}

\mypara{Definition of the quality score.}
The proposed video caption scorer is expected to evaluate a given video-caption pair by assigning a quality score that reflects their alignment degree.
A higher quality score means that the caption accurately describes more elements present in the video, with fewer omissions or errors.
To calculate the quality score, the scorer is instructed to generate five individual scores and calculate their average as the final quality score of the input caption.
Each of these scores ranges from 1 to 5 and focuses on a specific type of visual element: object, object feature, object action, camera movement, and background. 
If a particular type of visual elements is not involved in the video (e.g., a video without camera movement), a score of 0 is assigned, and this score is excluded from the calculation. 
This ensures that videos are not unfairly penalized for lacking certain elements, as they may be training-valuable on the remaining aspects.
Formally, the quality score of a synthesized video caption can be calculated as follows:
\[
\text{Quality Score} = \frac{\sum_{i=1}^{5} s_i \cdot \mathbb{I}(s_i \neq 0)}{\sum_{i=1}^{5} \mathbb{I}(s_i \neq 0)},
\]
where $s_i$ is the score for the $i$-th visual element type, $i \in \{1,2,3,4,5\}$ corresponds to: object, object feature, object action, camera movement, and background. 
$\mathbb{I}(s_i \neq 0)$ is an indicator function that equals 1 if $s_i \neq 0$ and otherwise 0.

\mypara{Structured annotation pipeline.}
The annotation procedure is structured as a series of five question-answering tasks since annotators report that roughly assigning a general and unstructured score is unclear and confusing, thereby significantly lowers annotation efficiency and quality.
For each task, the annotators are shown with a video, a synthetic caption that describe the video, and five questions about one type of fine-grained video-caption alignment.
Each question offers six answer options representing the aforementioned quality score, an integer ranging from 0 to 5.
To complete the annotation, annotators have to answer the five given questions considering the alignment degree of the video-caption pair on a specific type of visual elements.
Finally, the average score of these five questions is employed as the overall quality score of the synthesized caption.
In cases that the annotated video is low in imaging quality or descriptive value, such as video contains NSFW content or no meaningful content, we ask the annotators to tag these videos and drop them.
Notably, we also incorporate a rigorous quality control mechanism including user interface design, multi-turn trial annotation, multi-turn quality inspection and so on (see supplementary files).
In general, applying this structured annotation pipeline not only allows for a consistent and comprehensive yielding of human-aligned caption scoring data but also strikes a considerable balance between cost and efficiency.
For more details related to the annotated dataset statistics, please refer to the supplementary files.

\mypara{Annotated data statistics.}
To summarize, with videos sampled from OpenVid-1M and captions generated by the three ensembled base models, we annotate 6,534 video-caption pairs in total.
2,536 video-caption pairs are dropped because of the low quality of the paired video, with the remaining 4,008 pairs are successfully annotated with the five aforementioned structured quality scores.
Consequently, we format these metadata into 20,040 question-answering pairs to train our scorer model.

\subsubsection{Scorer Model Implementation}
\label{subsubsec:scorer_implementation}
We employ LoRA~\citep{hu2021lora} fine-tune to the VILA-v1.5-13B model on the annotated data, resulting in our scorer model.
During training, we finetune all the linear layers in the connector and LLM modules with the rank and alpha of LoRA set as 256 and 512, respectively.
The whole training is conducted on 8 NVIDIA A100 GPUs with a batch size of 16.
The learning rate warms up from 0 to 2e-5 in the first 3\% steps and applies a cosine learning rate scheduler.


\subsection{Ensemble Multi-models' Advantages and Human Preference}
We devise a three-stage training pipeline to implement the proposed ensembling synthetic and human preferenced training while meet common engineering need.
Specifically, we curate data employing the scorer-based selection policy with threshold, train Cockatiel-13B based on the data, and distill Cockatiel-8B from Cockatiel-13B.
More details can be found in the following paragraphs in \cref{subsubsec: pipeline}, while model implementations of Cockatiel captioner can be accessed in \cref{subsubsec:captioner_implementation}.

\subsubsection{Training Pipeline}
\label{subsubsec: pipeline}

\mypara{Human-aligned data curation.} 
Since VDCSCORE evaluates captions across five distinct perspectives, each with a specialized set of prompts, we directly employ these prompts to instruct the base models to generate captions focusing on these dimensions.
Initially, for each perspective, we sample 100k videos from OpenVid-1M~\citep{nan2024openvid-1m} and pair each video with perspective-centered captions generated by three base models, with each model producing exclusive one caption.
Then, we score each candidate caption and exclusively involve the one with the highest score for training if it exceeds the preset threshold.
Specifically, we first leverage our scorer model to assign quality scores to the captions.
The paired caption with the highest score is considered as the optimal candidate for that video.
Further, to introduce human preference and address cases where all three base models perform poorly on certain videos, we set a threshold of 3.5 on the quality score to filter out samples that are less human-preferable or low in caption quality.
After that, the remaining data of each perspective that meets our requirements is sampled to a fixed size, 20K, to balance the inter-dimension proportion.
Eventually, we curate a high-quality and human-preferable training dataset that contains 100K pairs between dimension-specific caption and videos, with 20K video-caption pair for each perspective.

\mypara{Ensemble synthesized training.}
Through the data curation, we obtain a training dataset that not only ensembles the strength of each base model but also ensures a high alignment in fine-grained video-caption and human preference.
Then, we finetune VILA-v1.5-13B on the curated data, which features strength-ensemble of the frontier VDC models and high alignment with human preference.
Notably, the large model capacity of the backboned VILA-v1.5-13B also guarantees the generalization of our model.

\mypara{Distilling 8B captioner model.}
For the ease of user usage and deployment, we further distill a smaller Cockatiel-8B model from the Cockatiel-13B, leveraging VILA-v1.5-8B as the foundation model.
Rather than directly fine-tuning the model on captions generated by Cockatiel-13B, we incorporate Cockatiel-13B as an additional base model.
In other words, the procedure of the distillation stage is similar to that in the ensemble training stage, except the model to be fine-tuned on is VILA-v1.5-8B and the integrated base models have an additional member, Cockatiel-13B.
Such procedure is approved because the former considers the situation where the Cockatiel-13B does not consistently outperform other base models.
Besides, compared to the latter procedure, we incur no additional costs beyond the cheap scoring of captions generated by Cockatiel-13B.

\subsubsection{Captioner Model Implementation}
\label{subsubsec:captioner_implementation}
Cockatiel-13B and Cockatiel-8B share identical hyper-parameter settings with the scorer model, which is described in \cref{subsubsec:scorer_implementation}.
As for their optimization processes, it can be formulated as follows:

For simplicity, we define the training sample $S$ for the captioner as a triplet, $S=(V,I,C)$, where $ V \in \mathbb{R}^{T \times H \times W \times 3} $ is the input video requiring captioning, with $T$ frames, each of height $H$ width $W$, and 3 color channels.
While $I$ is the textual instruction that prompts the captioner to generate a caption for the video and $ C = (C_1, C_2, \ldots, C_n) $ is the ground truth caption, which is the desired output, with $n$ tokens.

First, we uniformly sample 32 frames from the video $V$ and represent it as a frame sequence $V_s = \{v_1, v_2, \ldots, v_{32}\}$, where each $v_t \in \mathbb{R}^{H \times W \times 3}$ is a single frame.
The frame sequence $V_s$ is then fed into the Vision Transformer~(ViT) and projection module to extract the projected video feature  $F \in \mathbb{R}^{d \times 32}$ :
\begin{equation}
    F = \text{Proj}(\text{ViT}(V_s)).
\end{equation}
The functions $\text{ViT}(\cdot)$ and $\text{Proj}(\cdot)$ denote the process of encoding the frame sequence and mapping the ViT output to a fixed-dimensional feature $F$, respectively.

Next, the projected video feature $F$ is concatenated with the instruction $I$ and passed into the LLM to generate the caption. 
The optimization objective is the general auto-regressive loss function calculated only on the ground truth caption $C$~\citep{lagler2013gpt2, floridi2020gpt3}:
\begin{equation}
    L(\theta) = \sum_{i=1}^n \log p(C_i \mid F, I, C_{<i}; \theta),
\end{equation}
where $n$ is the length of the ground truth caption $C$, $C_{<i} = (C_1, C_2, \ldots, C_{i-1})$ denotes the token sequence of \( C \) preceding the \( i \)-th token, $\theta$ represents model parameter.


\section{Results \& Analysis} 

\subsection{Main Results}

\mypara{Quantitative results.}
\cref{tab:vdc_benchmark_results} presents the quantitative comparison results between Cockatiel-13B and baseline methods. 
Owing to our ensembling synthetic and human preferenced training, our model achieves new state-of-the-art performance on VDCSCORE exclusively in a considerable dimension-balanced way.
It also notes that our model is capable to generate dimension-comprehensive and human-preferred detailed video captions. 

\begin{figure*}[t]
    \centering
    \includegraphics[width=.95\textwidth]{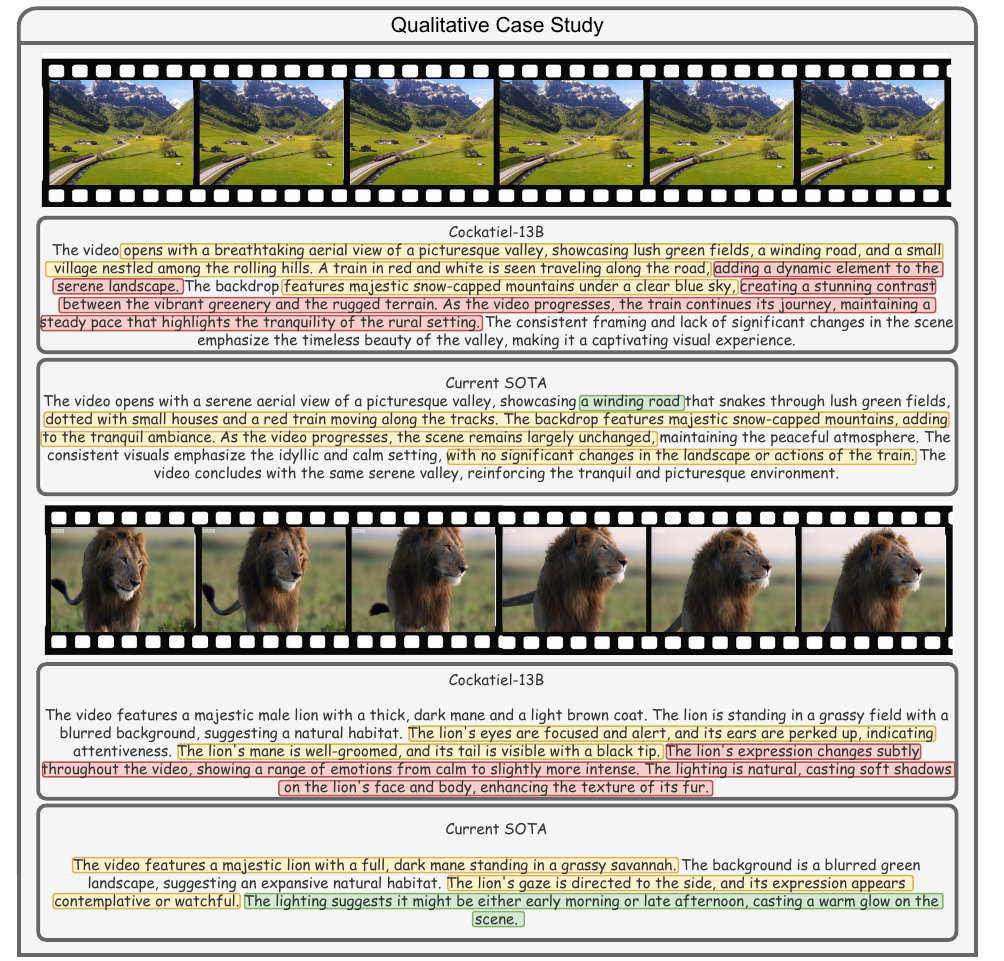}
    \vspace{-7pt}
    \caption{
    \textbf{Qualitative comparison between Cockatiel-13B and the current sota VDC models}. 
    For a detailed comparison between Cockatiel-13B and all leading VDC models, please refer to the supplementary files.
    The caption content that is exclusively captured by our model, captured by our model and other baselines, or misaligned with the detailed visual elements in the videos are emphasized using red, yellow and green backgrounds.
    }
    \label{fig:main_caption_case}
\end{figure*}

\mypara{Qualitative results.}
\cref{fig:main_caption_case} showcases the qualitative results generated by Cockatiel-13B and baseline methods, give a more direct and clearer comparison of them.
Through these cases, it is obvious that our model generates captions with better fine-grained video-caption alignment than baselines.
For instance, in the top case of \cref{fig:main_caption_case}, Cockatiel-13B exclusively captures the dynamic changes as the video progresses and differ the static part with the dynamic part of the video.
Furthermore, even the state-of-the-art VDC models that performs best on these cases generate content unaligned with input videos while Cockatiel-13B's generated captions demonstrate higher level of fine-grained video-caption alignment with no errors observed.

\begin{figure*}[]
\centering
\begin{subfigure}[b]{0.33\textwidth}
    \centering
    \small
    \begin{tikzpicture}
    \begin{axis}[
        width=5.5cm, height=4.5cm,
        xlabel={LoRA Rank},
        ylabel={Accuracy},
        xmin=0, xmax=10,
        ymin=43.2, ymax=44.0,
        xtick={0,2,4,6,8,10},
        xticklabels={0,64,128,256,512,Full},
        ytick={43.2,43.4,43.6,43.8,44.0},
        yticklabel style={/pgf/number format/fixed},
        axis lines=left,
        every axis plot/.append style={thick},
        nodes near coords, 
        every node near coord/.append style={font=\small, yshift=5pt}
    ]
        \addplot[
            color=black,
            mark=*,
            mark size=2pt
        ]
        coordinates {
            (2,  43.72)
            (4, 43.84)
            (6, 43.80)
            (8, 43.70)
            (10,43.43)
        };
    \end{axis}
    \end{tikzpicture}
\end{subfigure}
\hfill
\begin{subfigure}[b]{0.33\textwidth}
    \centering
    \begin{tikzpicture}
    \begin{axis}[
        width=5.5cm, height=4.5cm,
        xlabel={Dataset Size},
        xmin=0, xmax=10,
        ymin=42.8, ymax=44.0,
        xtick={0,2,4,6,8,10},
        xticklabels={0,10K,15K,20K,25K,30K},
        ytick={42.8,43.2,43.2,43.6,44.0},
        yticklabel style={/pgf/number format/fixed},
        axis lines=left,
        every axis plot/.append style={thick},
        nodes near coords, 
        every node near coord/.append style={font=\small, yshift=5pt}
    ]
        \addplot[
            color=black,
            mark=*,
            mark size=2pt
        ]
        coordinates {
            (2, 42.97)
            (4, 43.56)
            (6, 43.80)
            (8, 43.60)
            (10,43.74)
        };
    \end{axis}
    \end{tikzpicture}
\end{subfigure}
\hfill
\begin{subfigure}[b]{0.33\textwidth}
    \centering
    \begin{tikzpicture}
    \begin{axis}[
        width=5.5cm, height=4.5cm,
        xlabel={Threshold},
        xmin=0, xmax=6,
        ymin=43.3, ymax=44.0,
        xtick={0,2,4,6},
        xticklabels={0,2.5,3.0,3.5},
        ytick={43.2,43.4,43.6,43.8,44.0},
        yticklabel style={/pgf/number format/fixed},
        axis lines=left,
        every axis plot/.append style={thick},
        nodes near coords, 
        every node near coord/.append style={font=\small, yshift=5pt}
    ]
        \addplot[
            color=black,
            mark=*,
            mark size=2pt
        ]
        coordinates {
            (2, 43.43)
            (4, 43.56)
            (6, 43.80)
        };
    \end{axis}
    \end{tikzpicture}
\end{subfigure}
\vspace{-15pt}
\caption{
\textbf{Ablation studies on the LoRA rank (\textit{left}), training dataset size (\textit{middle}), and the quality score threshold (\textit{right})}.
%
For brevity, we report only the average accuracy on VDCSCORE; more comprehensive results are provided in the supplementary materials.
The hyper-parameter settings are consistent across all the ablation studies, except the ablated one. 
Specifically, the default settings are as follows: LoRA rank is set to 256, the training dataset size is 20k, and the threshold for the quality score is 3.5, the selection policy is the scorer-based selection policy with threshold on quality score.
    }
\label{fig:ablation}
\end{figure*}
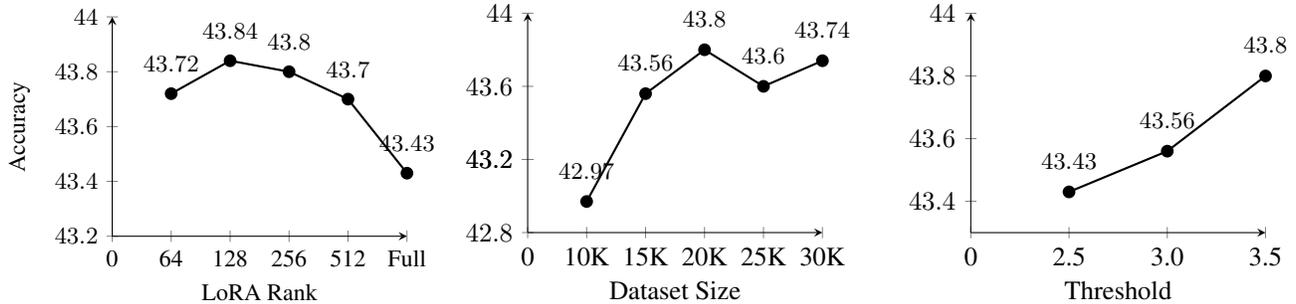

 \begin{figure}[]
    \centering
    \includegraphics[width=.9\linewidth]{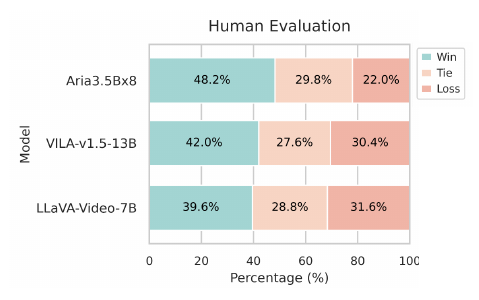}
    \vspace{-12pt}
    \caption{
    \textbf{Human evaluation results}. Our method, Cockatiel-13B, is obviously more human-preferred compared to baselines.
    }
    \label{fig:human_evaluation}
\end{figure}

\mypara{Human evaluation.}
To validate whether we successfully infuse Cockatiel-13B with human preferences and evaluate the generalization ability of our scorer on unseen captions, we conduct a pairwise human evaluation at instance level to compare Cockatiel-13B with three leading VDC models.
The results of human evaluation are summarized in \cref{fig:human_evaluation}.
Apparently, Cockatiel-13B is consistently voted as the most human-preferred VDC model, outperforming other baselines by a significant margin.
The results also evidence the effectiveness of our scorer and our method since they successfully align Cockatiel-13B with human preference.

\subsection{Ablation Studies}

\begin{table*}[t]
\centering
\small
\begin{tabular}{l|cccccc}
\toprule
\multicolumn{1}{c|}{Policy}                                                        & \begin{tabular}[c]{@{}c@{}}Camera\\ (Acc / Score)\end{tabular} & \begin{tabular}[c]{@{}c@{}}Short \\ (Acc / Score)\end{tabular} & \begin{tabular}[c]{@{}c@{}}Background\\ (Acc / Score)\end{tabular} & \begin{tabular}[c]{@{}c@{}}Main Object\\ (Acc / Score)\end{tabular} & \begin{tabular}[c]{@{}c@{}}Detailed\\ (Acc / Score)\end{tabular} & \begin{tabular}[c]{@{}c@{}}Average\\ (Acc / Score)\end{tabular} \\ 
\midrule
Random & 40.97/2.12 & 42.98/2.22 & 42.20/2.19 & 43.16/2.23 & 42.71/2.21 & 42.40/2.19 \\
VILA-zeroshot & 42.00/2.17 & 43.49/2.25 & 42.93/2.22 & 43.51/2.25 & 43.95/2.26 & 43.18/2.23 \\ \midrule
Ours w/~~ threshold  & 42.62/2.21 & 43.45/2.25 & 44.13/2.28 & 44.37/2.29 & 44.42/2.29 & 43.80/2.26 \\
Ours w/o threshold & 42.87/2.21 & 43.11/2.23 & 43.59/2.25 & 43.94/2.26 & 43.65/2.25 & 43.43/2.24 \\

\bottomrule
\end{tabular}
\vspace{-7pt}
\caption{
    \textbf{Ablation study results on selection policy.}
     ``Random'', ``VILA-zeroshot'', ``Ours w/ threshold'', ``Ours w/o threshold'' are abbreviations for the compared selection policies.
     Specifically, they represent random selection, ranking-based selection using VILA, and scorer-based selection with or without the threshold setting.
}
\label{tab:ablation_policy}
\end{table*}

\mypara{Ablation study on selection policy.} 
Selection policy determines how we select the most advantageous caption among its candidates, ensuring the effectiveness of our method.
Considering its importance, we devise various selection policies such as random selection and ranking-based selection using VILA, and the scorer-based selection policy with/without the threshold setting.
As depicted in \cref{tab:ablation_policy}, all selection policies enhance VDCSCORE performance, while the scorer-based one with threshold setting achieving the highest improvement, by an average gain of 1.84.
Notably, the scorer-based policy solely provides marginal advantages.
However, integrated with threshold setting, it can be further bootstrapped by 0.37 on VDCSCORE, demonstrating the superiority of this composition.

\mypara{Ablation study on training dataset size.}
To select a training dataset size that optimally balances cost and effectiveness while serving a practical training recipe for further research on VDC, we conduct ablation study on training dataset size.
The results are shown in \cref{fig:ablation}.
Initially, the average accuracy of VDCSCORE consistently improves as the training dataset size increases.
However, when the dataset size scales from 20K to 25K and 30K, the model performance reaches convergence with no notable improvement observed.
Surprisingly, the ablation study on training dataset size reveals that data scaling is not always useful in our method, which may be caused by the synthetic nature of our training data.
In conclusion, our findings indicate that 20K is an optimal training dataset size, and scaling beyond this size does not necessarily lead to better performance.

\mypara{Ablation study on threshold setting.}
In addition to the selection policy, we also employ a threshold on the caption quality score to ensure the training data quality.
To verify the effectiveness of our scorer and explore the optimal threshold setting, we conduct an ablation study on it.
As the results depicted in \cref{fig:ablation}, with the threshold increases from 2.5, the average accuracy of VDCSCORE improves significantly and monotonically, demonstrating the effectiveness of our scorer model.
Notably, the maximum threshold explored is 3.5, as insufficient data meets the criteria for thresholds of 4.0 or higher.
Combined with the ablation study on training dataset size, the results from the threshold study suggest that, for ensemble synthesized training in VDC, the quality of the training data has a more significant impact on model performance than its quantity.

\mypara{Ablation sutdy on LoRA rank.}
Increasing the LoRA rank allows more trainable parameters and incorporates more VDC-related knowledge into the base model, but it also introduce impairment on model generalization and general task performance.
As a consequence, we also investigate the impact of LoRA rank on VDC performance and summarize the results in \cref{fig:ablation}. 
Following common practices, throughout the study, the LoRA alpha is consistently set to twice the LoRA rank while other hyperparameters remain at their optimal settings.
The model achieves its best performance when the LoRA rank is set to 128 or 256.
However, the performance drops significantly when increasing the rank to 512 or applying full fine-tuning.
In consideration of accommodating more captioning knowledge, we set the LoRA rank as 256 instead of 128.

\section{Conclusions}
In this paper, we raise two challenges faced by existing VDC models, the imbalanced detailed video-caption alignment and the misalignment with human preference.
To alleviate these challenges, we develop Cockatiel, a three-stage training pipeline which assembles the advantages of multiple VDC models and human preferences by conducting ensembling synthetic and human preference training on VDC models based on our caption quality scorer.
It involving exclusively the candidate captions with the highest score for training if it exceeds the preset threshold. 
Further, we develop Cockatiel captioner series employing our training pipeline.
Extensive results verify the effectiveness of our method, as Cockatiel-13B demonstrates significant and dimension-balanced performance on VDCSCORE while achieving high alignment with human preferences.

{
    \small
    \bibliographystyle{ieeenat_fullname}
    \bibliography{main}
}

\clearpage
\appendix

\section{Appendix}
Here, we demonstrate the organization of the supplementary material to help our readers to quickly find the content they interest.

First, we introduce the possible limitations as well as applications of Cockatiel, in \cref{supp_sec:limitation} and \cref{supp_sec:application}.
The related works of this paper are introduced in \cref{supp_sec:related_work}.
Specifically, we introduce the works of video captioning and video Multimodal Large Language Models (video MLLMs) in \cref{supp_subsec:related_video_caption} and \cref{supp_subsec:related_video_mllm}.

After that, we introduce more details related to the annotation for the dataset of structured human preference score on video detailed captions in \cref{supp_sec:annotation}. 
The annotation involves a comprehensive quality control mechanism to ensure the quality, quantity and diversity of the annotation, which will be generally introduced in \cref{supp_subsec: anno_quality_control}.
As for details within the quality control mechanism, such as details related to the annotator selection and training, annotation user interface design and the prompts we used to instruct both the annotators and our scorer model, are introduced in \cref{supp_subsec: anno_human_resource}, \cref{supp_subsec:anno_user_interface}, \cref{supp_subsec:anno_prompt}, respectively.

Finally, we additionally give more experimental results and analysis in \cref{supp_sec:exp}. 
We exhibit the detailed results of every ablation study involved in theis paper, details related to human evaluation, and more qualitative case study in \cref{supp_subsec:exp_ablation}, \cref{supp_subsec:exp_human_evaluation}, and \cref{supp_subsec:exp_case}, respectively.

\section{Limitations}
\label{supp_sec:limitation}
\mypara{Scorer performance and base models' capabilities.} 
The potential of our method is restricted by two key factors: the performance of the scorer and the capabilities of the ensembled base models.
On the one hand, the capabilities of the base models define the upper-bound performance of our method. 
If the base models collectively perform poorly in specific aspects of VDC, we cannot generate high-quality synthesized data for them, and consequently, cannot infuse knowledge of those aspects into our captioner model.
On the other hand, the scorer determines how close we can approach the theoretical upper-bound performance.
The more accurate the scores assigned, the higher the quality of training data we can ultimately obtain.
Furthermore, like other works that utilize LLM, our methods are prone to encounter hallucination, where they generate content that appears plausible but is factually incorrect, fabricated, or misaligned with the input videos.
Despite these constraints, we believe our method is off to a strong start, serving as a competitive baseline for future works.
Our scorer is a powerful MLLM fine-tuned on a meticulously-annotated dataset of structured human preference scores for caption quality.
This ensures its effectiveness and generalization ability, which is further validated through ablation studies on selection policy and threshold setting.
As for the ensembled base models, we select the top-3 competitive models on VDCSCORE, which represent the frontier performance of existing VDC models.
Together, these components provide a solid foundation for our approach.

\mypara{Model capacity.}
Due to limited GPU resources, in this paper, we exclusively focus on VDC models that can be accommodated in a single GPU, specifically those with 34B parameters or fewer. 
As a result, the conclusions drawn in this paper remain untested and may not hold for VDC models larger than 34B, especially the findings related to LoRA rank and training dataset size.
That said, while the optimal hyper-parameter settings might vary for larger models, the effectiveness and efficiency of our framework are likely to experience only marginal changes. 
This is because our framework is not strictly dependent on the model capacity of the base models, but rather on the idea of ensemble synthesized training based on human-aligned caption quality score, which is scalable across different model sizes.

\mypara{Video length} 
Considering our captioner models are mainly trained on videos shorter than 60 seconds and without scene transitions, their performance is limited when handling longer videos or those containing scene transitions.
Nevertheless, it is still feasible to input such videos into our models and obtain captions reasonably aligned with them by increasing the number of input frame or employing the differential captioning technique introduced in \citet{chen2024sharegpt4video}.
Besides, since our captioner models are fine-tuned versions of VILA v1.5, they may inherit the capability to handle such videos from the base model. 
However, this inherited capability is not explicitly optimized in our current framework, leaving room for future improvements in addressing these challenges.

 \begin{figure*}[t]
    \centering
    \includegraphics[width=.9\textwidth]{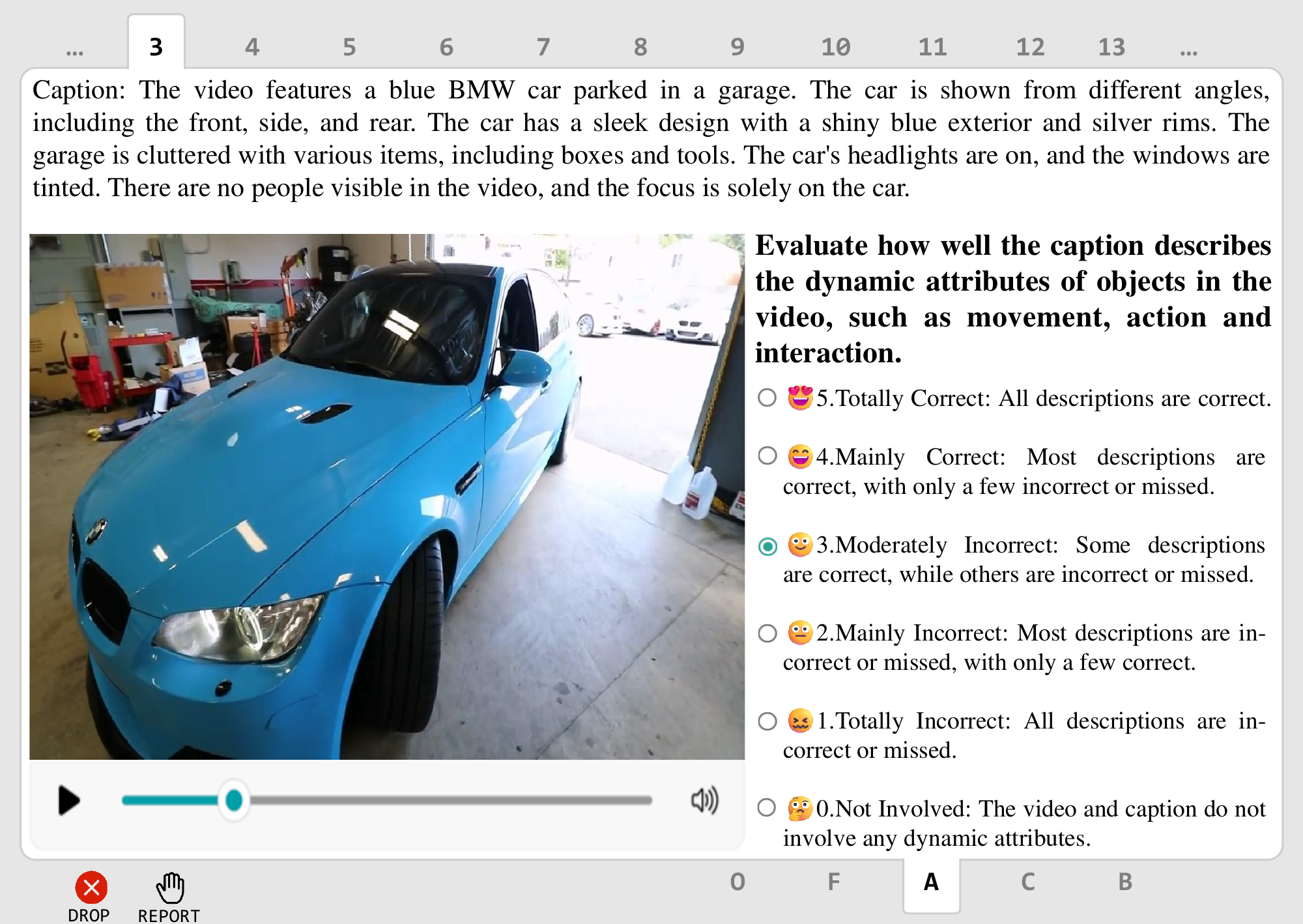}
    \caption{
    A snapshot of our annotation user interface, where the whole annotation procedure is carried on. 
    In each annotation task, the annotators are present with a video on the left, a caption supposed to align with it on the top, and a question related to the detailed video-caption alignment on the right of the user interface. 
    ``O'', ``F'', ``A'', ``C'', ``B'' are abbreviations for ``Object'', ``object Feature'', ``object Action'', ``Camera'' and ``Background'', respectively.
    Notably, our user interface is highly praised by our annotators for its user-friendly and intuitive design, which ensures both the quality and quantity of the annotated data.
    }
    \label{fig:user_interface}
\end{figure*}

\section{Applications}
\label{supp_sec:application}

Considering the fundamental effect of our method and video detailed captioning task, we believe our works can be applied to various scenarios, such as evaluating video detailed captions, and serve as synthetic data generator for the training of other video understanding models.

\mypara{Evaluating video detailed captions.} 
The yield of our scorer model, the quality score, is closely aligned with human preferences on evaluating model-generated video detailed captions, which serves as an alternative for human evaluation, GPT4-o based evaluations, and other costly evaluation methods.
Besides, the scorer component can be used for filtering low-quality caption data. Since the score assigned by our scorer has clear and practical meanings, we suggest users set a threshold greater than 3, which indicates that almost half of the visual elements involved in the video is missed or not correctly described.
Meanwhile, setting a too high threshold may introduce astonishing computation costs.


\mypara{Video detailed captioning for VQA, fine-grained video understanding.}
Cockatiel-13B achieves new state-of-the-art performance on VDCSCORE, demonstrating considerably competitive performance on video detailed captioning and significant potential for providing structured detailed video captions.
Thus, by feeding it with videos, our method can continuously generate high-quality captions that can be used for synthesized training.
As employing synthetic video detailed caption data for training is quite common in works about video question-answering and video-understanding~\citep{li2024llava-ov, kim2024generalizing_vqa}.

\section{Related Work}
\label{supp_sec:related_work}

\subsection{Video Captioning}
\label{supp_subsec:related_video_caption}

Video captioning aims to understand the input videos and generate captions aligned with its visual content, dynamic semantics, and temporal information, bridging the modality gap between text and video.
Typically, existing video captioning datasets, such as MSRVTT~\citep{xu2016msrvtt} and MSVD~\citep{chen2011msvd}, only pair each video with one or two sentences, which is quite short and thereby insufficient to describe most of the semantic and temporal information involved in videos.
Meanwhile, the emergence of multimodal large language models and text-to-video diffusion models has catalyzed transformative progress in video-based AI systems and high-demand on descriptive video captioning tools.
Motivated by this, researchers turn to build state-of-the-art video detailed captioning models to bootstrap the development of video understanding and generation~\citep{chen2024panda-70m, yang2024vript, ju2024miradata, chen2024sharegpt4video, zhang2024llava-vid, chai2024auroracap}.

Unfortunately, developing a competitive video detailed captioning model is fraught with various challenges, including the scarcity of high-quality training data and the difficulty of achieving balanced alignment between video content and captions.
These challenges are exacerbated by the inherent complexity of video detailed captioning.
Specifically, video detailed captioning is demanding even for humans, as it requires capturing every visual element and its dynamics across the temporal dimension.
To address the data scarcity challenge, \citet{chen2024panda-70m} propose to train a video retrieval model to identify the best-match candidates for each captioned video. 
Furthermore, \citet{yang2024vript} and \citet{chen2024sharegpt4video} focus on training VDC models using synthesized data distilled from GPT-4V~\citep{achiam2023gpt4} and GPT-4o~\citep{openai2024gpt4o}.
\citet{chai2024auroracap} propose the token merging technique to lower the computation requirement of video detailed captioning.

In this paper, we propose to ensemble the data generated by open-source models with respective advantages and filter the data using human-aligned caption scoring models.
In this way, our method successfully tackles both of the two aforementioned challenges.
By leveraging data generated by open-source models and employing a rigorous data filtering procedure, the training data of VDC can be both cost-effective and high-quality, thus sufficiently mitigating the first challenge.
As for the second challenge of imbalanced alignment, we alleviate it by ensembling data generated by models with diverse specialties, as the fine-tuned models exhibit improvements across all of the evaluated dimensions on VDCSCORE.
Notably, the cost of our approach is substantially lower than distilling data from commercial-level models, making it more scalable and accessible to research groups with limited budgets.

\subsection{Video Multimodal Large Language Models~(Video MLLMs)}
\label{supp_subsec:related_video_mllm}

Recent advancements in multimodal large language models~(MLLMs) have shown impressive performance in general image-text dialogue tasks, along with significant potential for extension to the video modality.
Inspired by this progress, researchers begin to study video multimodal large language models~(Video MLLMs), where MLLMs are enhanced to support the processing of text, images, videos, and any combination thereof.
The classic architecture of MLLM consists of three main components: a visual encoder, a connector, and a large language model.
The visual encoder first extracts features from the input image or video frames. 
Then, these features are passed through the connector, which projects them into the semantic space of the large language model. 
Finally, the projected visual features are concatenated with the embeddings of the input text and fed into the large language model for further processing.
For video inputs, videos are treated as sequences of ordered frames. 
Each frame is independently encoded by the vision encoder, and the resulting frame features are concatenated to form a frame-level representation.

As pioneer in this area, Video-ChatGPT~\citep{maaz2023video-chatgpt} proposes to concatenate the frame-level features with their pooled version to derive a comprehensive video-level feature.
While Video-LLaMA~\citep{zhang2023video-llama} integrating video along with its corresponding audio signal to the architecture of Video MLLMs by introducing video Q-Former and audio Q-Former.
integrating both video and corresponding audio signals into the MLLM architecture through the introduction of video Q-Former and audio Q-Former modules.
To address the problems of data scarcity and task diversity, \citet{zhang2024llava-vid} curates a large-scale video instruction-tuning dataset, LLaVA-Video-178K, which covers three common tasks in video-text dialogue. 
VideoChat2~\citep{li2024videochat2} transfers static image tasks into their dynamic video version and conduct progressive multi-modal training with diverse instruction-tuning data to establish robust video MLLM baselines.
Recently, to further scale training data and amplify applicability, the latest version of popular MLLM series, such as LLaVA-OneVision~\citep{li2024llava-ov}, NVILA~\citep{liu2024nvila}, Qwen2-VL~\citep{wang2024qwen2-vl}, and InternVL2.5~\cite{chen2024internvl2_5}, have incorporated video understanding capabilities into their frameworks.
These models have shown promising performance across a variety of video understanding tasks, marking a significant step forward in the field.

\section{More Annotation Details}
\label{supp_sec:annotation}

\begin{figure*}[t]
    \centering
    \includegraphics[width=.8\textwidth]{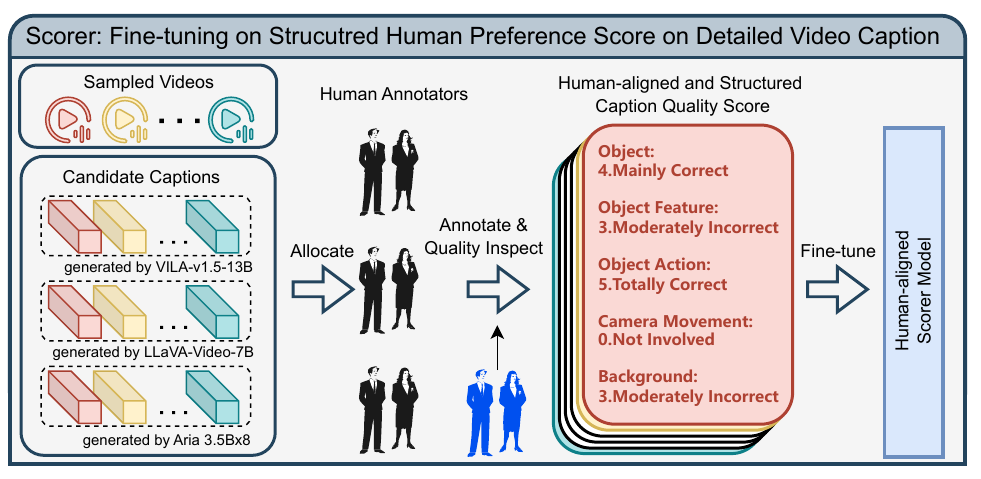}
    \vspace{-7pt}
    \caption{
    The training pipeline of our scorer.
    Our selection policy is critical as its performance determines which knowledge and strengths is ensembled into Cockatiel-13B. 
    As a consequence, since our aim is to infuse Cockatiel-13B all the strengths of the base models and human preferences, we devise a selection policy with threshold setting based on our human-aligned caption quality scorer.
    To obtain our caption quality scorer, we need human-annotated data on it or off-the-shelf models.
    Since no publicly available dataset nor model suits this need, we build them on our own, as demonstrated in this figure.
    }
    \label{fig:scorer_pipeline}
\end{figure*}

\subsection{Annotation Quality Control}
\label{supp_subsec: anno_quality_control}

Aside from the previously mentioned structured annotation pipeline, we also employ a rigorous quality control mechanism to guarantee the quantity, quality and efficiency of the annotation.
Specifically, the quality control mechanism involves user interface design design, annotator training, multi-turn trial annotation, as well as multi-turn quality inspection.
The entire annotation process is carried out on a specially designed user interface, which enables us to regulate annotator behavior and supports the quality control mechanism.
Before the formal annotation, we write all the details related to the goal, instruction and principle of the annotation in an annotation guidelines, and train our annotators to understand them.
Moreover, we have an official meeting with them to explain the guidelines to them and let them to raise questions on details they don't understand.
Before the formal annotation begins, we prepare a detailed annotation guideline that outlines the goals, instructions, and principles of the annotation.
Then, we hold official meetings to explain the guidelines in detail and address any questions or uncertainties raised by the annotators.
After that, we conduct a multi-turn trial annotation to optimize the annotation procedure and ensure that annotators adhere to the guidelines, maintaining high intra- and inter-annotator consistency.
During the annotation, we conduct 8 turns of random sampling quality inspection over the present annotated samples to guarantee ongoing accuracy and consistency. 
Each inspection involves a detailed review of the annotations by experts, who verify whether the annotations align with the guidelines and meet the required quality standards. 
If discrepancies or errors are identified, we feedback to the corresponding annotators and take corrective measures immediately.
This iterative process not only helps maintain high annotation quality throughout the project but also allows us to identify and address potential issues early, ensuring the reliability of the final annotated dataset.

\subsection{Annotator Selection and Training}
\label{supp_subsec: anno_human_resource}

The accuracy and reliability of annotated data are significantly depended on the capabilities of the human annotators involved in the annotation process.
Particularly, in our cases, our task of structured quality scoring on detailed video captions with human preferences are quite demanding on annotators.
Since it requires the human annotators to score abstract integers on detailed video captions and hold a consistent and appropriate annotation standard based on the combination of their subjective and objective cognition.
As a consequence, at the beginning of the annotation, We first conduct annotator selection to build an appropriate and unbiased annotation team, and train this annotation team with our meticulously prepared annotation guidelines.
For annotator selection, we let the candidates to accomplish a test concentrating on 10 factors, domain expertise, resistance to disturbing content, attention to detail, communication skills, reliability, cultural and linguistic competence, technical skills, ethical considerations, describing ability, and motivation.
Notably, since the videos and their paired captions may contain uncomfortable and inappropriate content, the candidates are notified with this inconvenience before the test.
Only those agreed with this inconvenience are eligible to participate in the test, and they are welcome to withdraw at any time if they choose to do so.
Based on the test results and candidate backgrounds, We try our best to ensure that the selected annotators are well-balanced in background and have a generally competitive abilities of the 10 mentioned factors.
To summarize, our annotation team includes 10 annotators carefully selected from 47 candidates, 5 males and 5 females, all have a bachelor's degree.
We interview the annotators and ensure they are adequate for the annotation.

\subsection{Annotation User Interface Design}
\label{supp_subsec:anno_user_interface}

\cref{fig:user_interface} illustrates the annotation user interface designed by us.
With this specialized user interface, we are able to establish sound communication mechanism, regulate annotator behavior, and support various quality control measures.
Here, we introduce details about the interaction involved within the annotation user interface.

At the top of the user interface,a series of task IDs are displayed as serial numbers.
The ongoing one is emphasized with bold number and white background, while others are shown with grey background and task IDs with no bold.
By clicking on the serial number, the user interface will navigate to the corresponding annotation task, showing the annotation content requires labeling.
To move to earlier or later tasks, the annotators can click on the omission of both ends to let the user interface display the task ID the annotators looking for.
The one on the left end will display the ID of earlier tasks, the one on the right will display the ID of later tasks.

Below the serial numbers is the main panel which consists of three components: the input video  with a video player component on the left, a caption expected to align with the vide on the top, and a question with six options on its right.
Annotators are supposed to watch the video thoroughly, read the caption carefully, and assess the alignment between them based on the question. 
Eventually, they select the most appropriate option according to the annotation guideline and their judgment to complete the annotation of the question.

As we mentioned above, each annotation task contains five questions that rigorously assess the degree of alignment between the caption and a specific type of visual elements involved in the video.
Specifically, the five types of visual elements are object, object feature, object action, camera movement and background, represented as the alphabet on the right bottom of the user interface, ``O'', ``F'', ``A'', ``C'', ``B'', respectively.
Similar to the serial numbers on the top, the alphabet that represents the ongoing question is highlighted with bold number and white background, while others are no bold and with gray background.
Users can jump to the panel related to a question by clicking on its corresponding alphabet.
Notably, if all the questions related to the present annotation task are accomplished, the user interface will automatically jump to the first question of the next task.
Otherwise, it will jump to one question of the present task that have not annotated yet.
Furthermore, on the bottom of the user interface, there are two buttons, one shaped like a red cross and the other shaped like a raised human hand, accompanied with the text describing their effect.
The red cross button is used to drop the data if the video is low in quality, such as videos poor in imaging quality, contains NSFW content, or contains incomprehensible material.
If the annotators are unsure about their selection on certain data, they can use the raise hand button to report to us by writing descriptions about their problem.

In general, our user interface is human-friendly and easy-to-use, which ensures efficient and accurate annotation while maintaining high-quality standards throughout the annotation process.

\subsection{Annotation Instruction Template}
\label{supp_subsec:anno_prompt}

Here, we present the prompts used to instruct our scorer model to generate structured quality scores for detailed video captions. 
These prompts can be categorized into two types.
First, we employ five task-specific prompts, each designed to evaluate the alignment degree between the caption and a specific visual element in the paired video: object, object feature, object action, camera movement, and background.
Second, these task-specific prompts share a system prompt, which provides overarching instructions to the scorer, clarifying the aim and expected behavior of the task.
Below are the demonstration of the system prompt and the five task-specific prompts (with input video and caption details omitted for brevity):

\begin{tcolorbox}
\textbf{[System]}

You are an expert in evaluating video captions, specifically focusing on how effectively they align with and describe the content of the videos.
\end{tcolorbox}

\begin{tcolorbox}
\textbf{[Object]}

Evaluate how well the caption describes the objects in the video.
Use the following scoring options:

0.Not Involved: The video and the caption do not involve any objects.

1.Totally Incorrect: All descriptions are incorrect or missed. 

2.Mainly Incorrect: Most descriptions are incorrect or missed, with only a few correct.

3.Moderately Incorrect: Some descriptions are correct, while others are incorrect or missed.

4.Mainly Correct: Most descriptions are correct, with only a few incorrect or missed.

5.Totally Correct: All descriptions are correct
\end{tcolorbox}
\begin{tcolorbox}
\textbf{[Object Feature]}

Evaluate how well the caption describes the static attributes of objects in the video (e.g., color, shape, size, and texture). Use the following scoring options:

0.Not Involved: The video and caption do not involve any static attributes.

1.Totally Incorrect: All descriptions are incorrect or missed .

2.Mainly Incorrect: Most descriptions are incorrect or missed, with only a few correct.

3.Moderately Incorrect: Some descriptions are correct, while others are incorrect or missed.

4.Mainly Correct: Most descriptions are correct, with only a few incorrect or missed.

5.Totally Correct: All descriptions are correct.

\end{tcolorbox}

\begin{tcolorbox}
\textbf{[Object Action]}

Evaluate how well the caption describes the dynamic attributes of objects in the video, such as movement, action and interaction. Use the following scoring options:

0.Not Involved: The video and caption do not involve any dynamic attributes.

1.Totally Incorrect: All descriptions are incorrect or missed.

2.Mainly Incorrect: Most descriptions are incorrect or missed, with only a few correct.

3.Moderately Incorrect: Some descriptions are correct, while others are incorrect or missed.

4.Mainly Correct: Most descriptions are correct, with only a few incorrect or missed.

5.Totally Correct: All descriptions are correct.

\end{tcolorbox}

\begin{tcolorbox}
\textbf{[Camera]}

Evaluate how well the caption describes the camera movement in the video, including moves, pans, tilts, and zooms. Use the following scoring options:

0.Not Involved: The video and caption do not involve any camera movements.

1.Totally Incorrect: All descriptions are incorrect or missed.

2.Mainly Incorrect: Most descriptions are incorrect or missed, with only a few correct.

3.Moderately Incorrect: Some descriptions are correct, while others are incorrect or missed.

4.Mainly Correct: Most descriptions are correct, with only a few incorrect or missed.

5.Totally Correct: All descriptions are correct.
\end{tcolorbox}

\begin{tcolorbox}
\textbf{[Background]}

Evaluate how well the caption describes the background (such as setting and context) in the video. Use the following scoring options:

0.Not Involved: The video and caption do not involve any background elements.

1.Totally Incorrect: All descriptions are incorrect or missed.

2.Mainly Incorrect: Most descriptions are incorrect or missed, with only a few correct.

3.Moderately Incorrect: Some descriptions are correct, while others are incorrect or missed.

4.Mainly Correct: Most descriptions are correct, with only a few incorrect or missed.

5.Totally Correct: All descriptions are correct.
\end{tcolorbox}

\section{More Experiment Results}
\label{supp_sec:exp}

\subsection{More Ablation Study Results}
\label{supp_subsec:exp_ablation}
In this subsection, we demonstrate more detailed results of the ablation study on training dataset size, LoRA rank. We also illustrate the ablation study and analysis on model distillation.

\mypara{Ablation study on model distillation.}
Typically, a simple yet efficient way to distill Cockatiel-8B from Cockatiel-13B is training the Cockatiel-8B on video detailed captions generated by Cockatiel-13B.
However, in our case, we already possess 100K structured detailed video captions generated by three leading VDC models and accompanied by caption quality scores assigned by our human-aligned caption scorer.
As a consequence, compared the this direct approach, there is no additional cost associated with involving both three base models and Cockatiel-13B for ensemble synthetic training but the cheap scoring on Cockatiel-13B generated captions. 
To verify the effectiveness of this design, we conduct an ablation study on the distilled models, the results of which are summarized in \cref{tab:ablation_distillation}.
As can be concluded, conducting another round of ensemble synthesized training on Cockatiel-13B and the three base models yields the most competitive results on VDCSCORE.
In addition, the gain of assembling the advantages of multiple base models is saliently significant.
Even when training solely on the most competitive model on VDCSCORE, Cockatiel-13B produces VDC models that underperform compared to simply ensembling the three base models.

\subsection{More Details about Human Evaluation}
\label{supp_subsec:exp_human_evaluation}
Here, we give a full demonstration of the conducted human evaluation.
To verify the alignment with human preferences of Cockatiel-13B, we conduct a pair-by-pair human evaluation was conducted at the case level.
Specifically, we compare Cockatiel-13B with VILA-v.15-13B, LLaVA-Video-7B and Aria3.5Bx8, since they are the top-3 competitive VDC models and base models of Cockatiel-13B.

\mypara{Data Preparation}
To prepare the data utilized for human evaluation, we first sample another 500 videos for detailed video captioning, except for those used for training.
Then, we let each compared models to caption the sampled 500 videos.
Finally, we make pairs between video, a caption generated by Cockatiel-13B that describes the video, and another one generated by one of the three compared alternatives.
We sample 500 videos and involve 3 baseline methods, VILA-v1.5-13B, LLaVA-Video-7B and Aria3.5Bx8 for the human evaluation, yielding totally 1,500 pairs to annotate.
To uniformly distribute the annotation tasks, we select 9 of 10 annotators from our annotation team and every is assigned 500 pairs of evaluation tasks.
The left one annotator evaluate whether the sampled videos have describing value and are suitable for the human evaluation.
In this way, the evaluation result of each caption pair has three annotations, we choose the option selected most as the evaluation result of this specific pair. 

\mypara{Evaluation Procedure.}
To conduct the human evaluation effectively and efficiently, we derive an easy-to-understand human evaluation user interface from that of our scoring human annotation for these human evaluators.
The difference between the two user interface are the user interface of human evaluation only choose a caption that they prefer.
Each time, users are provided with a video and two captions describing the video, one generated by Cockatiel-13B and the other one generated by one of the three baseline methods, VILA-v1.5-13B, Aria3.5Bx8 and LLaVA-Video-7B.
To accomplish the annotation, users have to read these two captions and choose the one that clearly aligns better with their general and personal preference.
If users find it quite difficult to find the one from the given two caption that they prefer, they can mark this pair as ``Not Distinguishable'' and we recognize this as a tie between the two evaluated models.
Notably, since we already checked the videos to be reviewed before, the ``DROP'' button in the user interface is set but not used during the human evaluation.
In cases that evaluators are confused about some details of a certain assigned evaluation task, they can utilize the raise hand button to ask the help of the organizers.
Thanks to our adequate user training, reasonable evaluation procedure and friendly user interface, the raise hand button is set but not used in our evaluation.
Notably, the captions are disordered, thus users are blind to the models that generates the comparing captions at the time.

\subsection{More Qualitative Results}
\label{supp_subsec:exp_case}

In the following pages, we give more detailed version of the qualitative analysis conducted in the main content, containing the complete comparison between the captions generated by Cockatiel-13B and three baseline methods, also the current top-3 VDC models according to the benchmark results on VDCSCORE.
Besides, we include four additional cases to further holistically demonstrate the effectiveness of our method and the superior performance of Cockatiel-13B.

\clearpage

\begin{table*}[t]
\centering
\begin{tabular}{c|cccccc}
\toprule
\multicolumn{1}{c|}{LoRA Rank}                                                        & \begin{tabular}[c]{@{}c@{}}Camera\\ (Acc / Score)\end{tabular} & \begin{tabular}[c]{@{}c@{}}Short \\ (Acc / Score)\end{tabular} & \begin{tabular}[c]{@{}c@{}}Background\\ (Acc / Score)\end{tabular} & \begin{tabular}[c]{@{}c@{}}Main Object\\ (Acc / Score)\end{tabular} & \begin{tabular}[c]{@{}c@{}}Detailed\\ (Acc / Score)\end{tabular} & \begin{tabular}[c]{@{}c@{}}Average\\ (Acc / Score)\end{tabular} \\ 
\midrule
64   & 42.80/2.22 & 43.62/2.25 & 43.99/2.27 & 44.14/2.28 & 44.05/2.27 & 43.72/2.26 \\
128  & 43.02/2.22 & 43.47/2.24 & 44.19/2.27 & 44.44/2.29 & 44.10/2.27 & 43.84/2.26 \\
256  & 42.62/2.21 & 43.45/2.25 & 44.13/2.28 & 44.37/2.29 & 44.42/2.29 & 43.80/2.26 \\
512  & 43.70/2.21 & 43.36/2.24 & 43.64/2.25 & 43.93/2.27 & 43.88/2.27 & 43.70/2.25 \\
Full & 42.60/2.20 & 43.29/2.24 &43.86/2.26 & 43.72/2.26 & 43.69/2.26 &43.43/2.24 \\
\bottomrule
\end{tabular}
\caption{Ablation study results on LoRA rank.``Full'' is the abbreviation for ``Full Fine-tuning''.}
\label{tab:ablation_lora_rank}
\end{table*}

\begin{table*}[t]
\centering
\begin{tabular}{c|cccccc}
\toprule
\multicolumn{1}{c|}{Dataset Size}                                                        & \begin{tabular}[c]{@{}c@{}}Camera\\ (Acc / Score)\end{tabular} & \begin{tabular}[c]{@{}c@{}}Short \\ (Acc / Score)\end{tabular} & \begin{tabular}[c]{@{}c@{}}Background\\ (Acc / Score)\end{tabular} & \begin{tabular}[c]{@{}c@{}}Main Object\\ (Acc / Score)\end{tabular} & \begin{tabular}[c]{@{}c@{}}Detailed\\ (Acc / Score)\end{tabular} & \begin{tabular}[c]{@{}c@{}}Average\\ (Acc / Score)\end{tabular} \\ 
\midrule
10k & 41.49/2.15 & 43.17/2.23 & 42.95/2.22 & 43.52/2.25 & 43.71/2.26 & 42.97/2.22 \\
15k & 42.55/2.20 & 43.67/2.26 & 43.66/2.26 & 43.95/2.27 & 43.95/2.27 & 43.56/2.25 \\
20k & 42.62/2.21 & 43.45/2.25 & 44.13/2.28 & 44.37/2.29 & 44.42/2.29 & 43.80/2.26 \\
25k & 42.46/2.20 & 43.51/2.24 & 43.85/2.26 & 43.90/2.27 & 44.28/2.28 & 43.60/2.25 \\
30k & 42.62/2.21 & 43.41/2.24 & 44.27/2.28 & 44.10/2.28 & 44.34/2.28 & 43.74/2.26 \\
\bottomrule
\end{tabular}
\caption{Ablation study results on training dataset size.}
\label{tab:ablation_datasize}
\end{table*}

\begin{table*}[t]
\centering
\begin{tabular}{c|cccccc}
\toprule
\multicolumn{1}{c|}{Threshold}                                                        & \begin{tabular}[c]{@{}c@{}}Camera\\ (Acc / Score)\end{tabular} & \begin{tabular}[c]{@{}c@{}}Short \\ (Acc / Score)\end{tabular} & \begin{tabular}[c]{@{}c@{}}Background\\ (Acc / Score)\end{tabular} & \begin{tabular}[c]{@{}c@{}}Main Object\\ (Acc / Score)\end{tabular} & \begin{tabular}[c]{@{}c@{}}Detailed\\ (Acc / Score)\end{tabular} & \begin{tabular}[c]{@{}c@{}}Average\\ (Acc / Score)\end{tabular} \\ 
\midrule
2.5  & 42.39/2.19 & 43.32/2.24 & 43.54/2.25 & 44.06/2.27 & 43.86/2.26 & 43.43/2.24 \\
3.0  & 42.80/2.22 & 43.25/2.24 & 43.85/2.26 & 43.86/2.27 & 44.06/2.26 & 43.56/2.25 \\
3.5  & 42.62/2.21 & 43.45/2.25 & 44.13/2.28 & 44.37/2.29 & 44.42/2.29 & 43.80/2.26 \\
\bottomrule
\end{tabular}
\caption{Ablation study results on quality score threshold.}

\label{tab:ablation_threshold}
\end{table*}

\begin{table*}[t]
\centering
\begin{tabular}{c|cccccc}
\toprule
\multicolumn{1}{c|}{Models}                                                        & \begin{tabular}[c]{@{}c@{}}Camera\\ (Acc / Score)\end{tabular} & \begin{tabular}[c]{@{}c@{}}Short \\ (Acc / Score)\end{tabular} & \begin{tabular}[c]{@{}c@{}}Background\\ (Acc / Score)\end{tabular} & \begin{tabular}[c]{@{}c@{}}Main Object\\ (Acc / Score)\end{tabular} & \begin{tabular}[c]{@{}c@{}}Detailed\\ (Acc / Score)\end{tabular} & \begin{tabular}[c]{@{}c@{}}Average\\ (Acc / Score)\end{tabular} \\ 
\midrule
Ours & 41.75/2.17 & 43.45/2.24 & 43.54/2.25 & 43.91/2.21 & 43.62/2.25 & 43.25/2.22 \\
Base & 42.23/2.19 & 43.77/2.26 & 43.63/2.25 & 43.76/2.26 & 44.04/2.27 & 43.49/2.25 \\ \midrule
Ours + Base  & 42.25/2.19 & 44.01/2.27 & 43.89/2.26 & 43.85/2.26 & 44.00/2.27 & 43.60/2.25 \\

\bottomrule
\end{tabular}
\vspace{-7pt}
\caption{
Ablation study results on model distillation. ``Ours'' and ``Base'' indicates Cockatiel-13B and its ensembled base models, respectively. }
\label{tab:ablation_distillation}
\end{table*}

\begin{figure*}[h!]
    \centering
    \includegraphics[width=.9\textwidth]{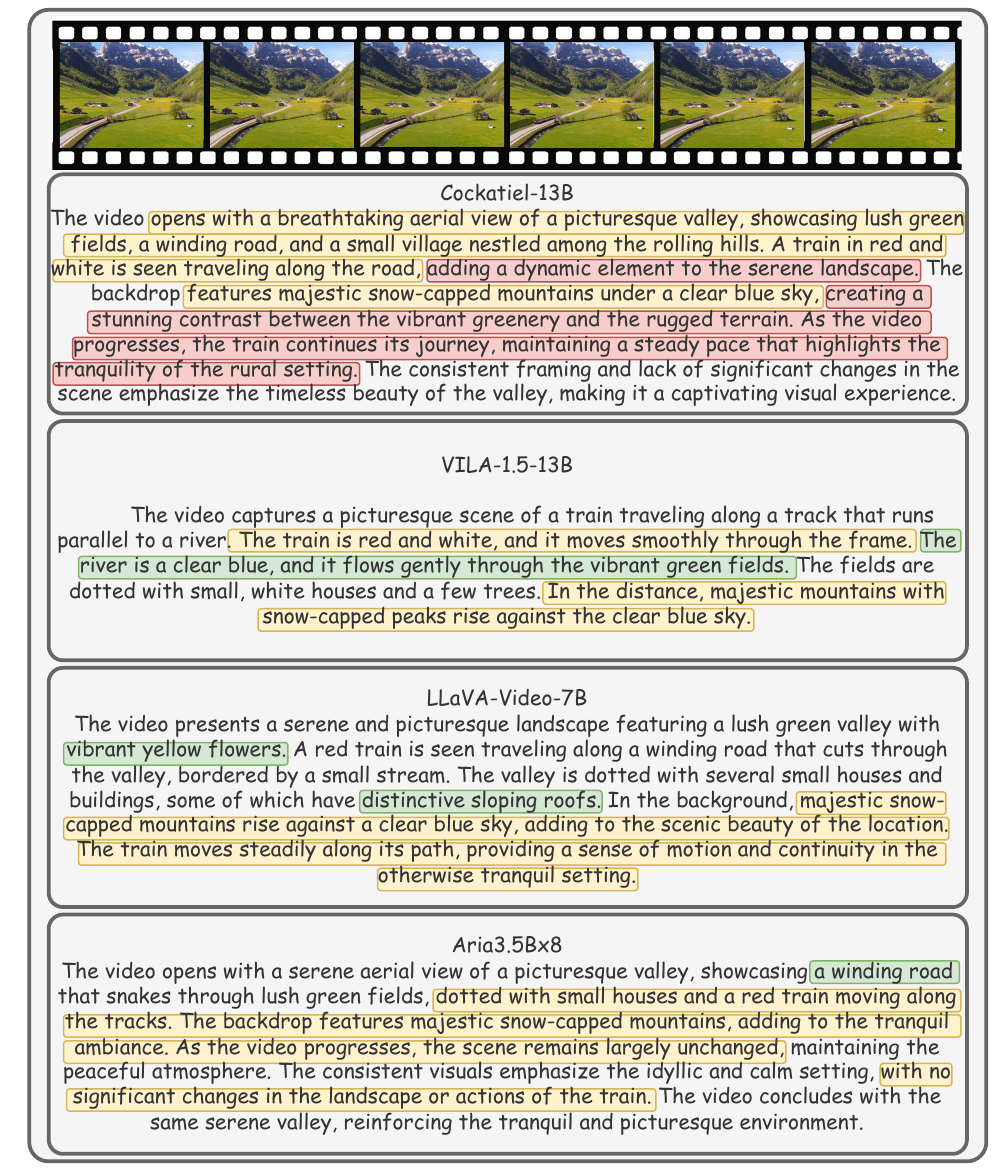}
    \caption{
    The first qualitative case compared with three base models in the main content. 
    The caption content that is exclusively captured by our model, captured by our model and other baselines, or misaligned with the detailed visual elements in the videos are emphasized using red, yellow and green backgrounds.
    }
\end{figure*}

\begin{figure*}
    \centering
    \includegraphics[width=.9\textwidth]{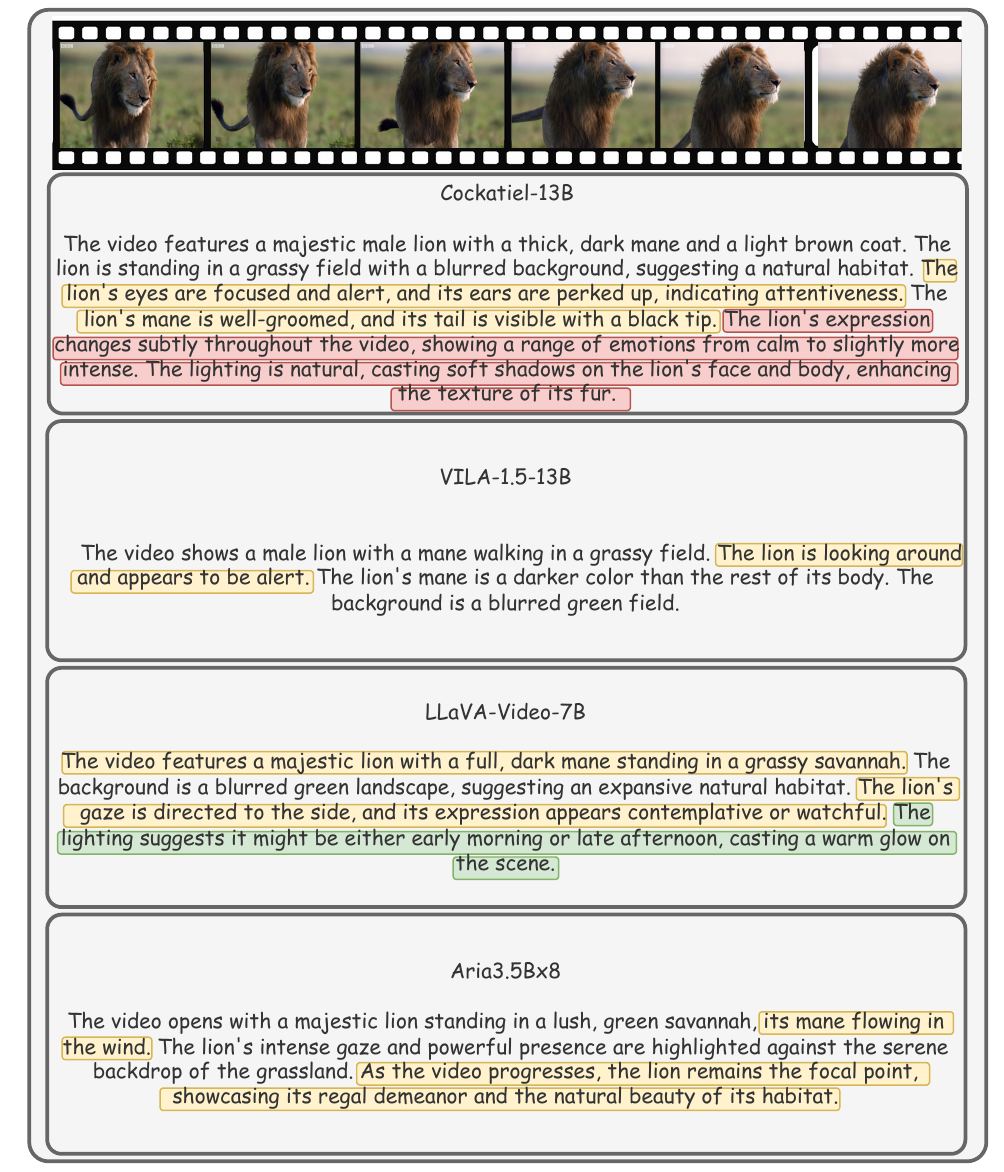}
    \caption{
    The second qualitative case compared with three base models in the main content. 
    The caption content that is exclusively captured by our model, captured by our model and other baselines, or misaligned with the detailed visual elements in the videos are emphasized using red, yellow and green backgrounds.
    }
\end{figure*}

\begin{figure*}[t]
\centering
\includegraphics[width=.9\textwidth]{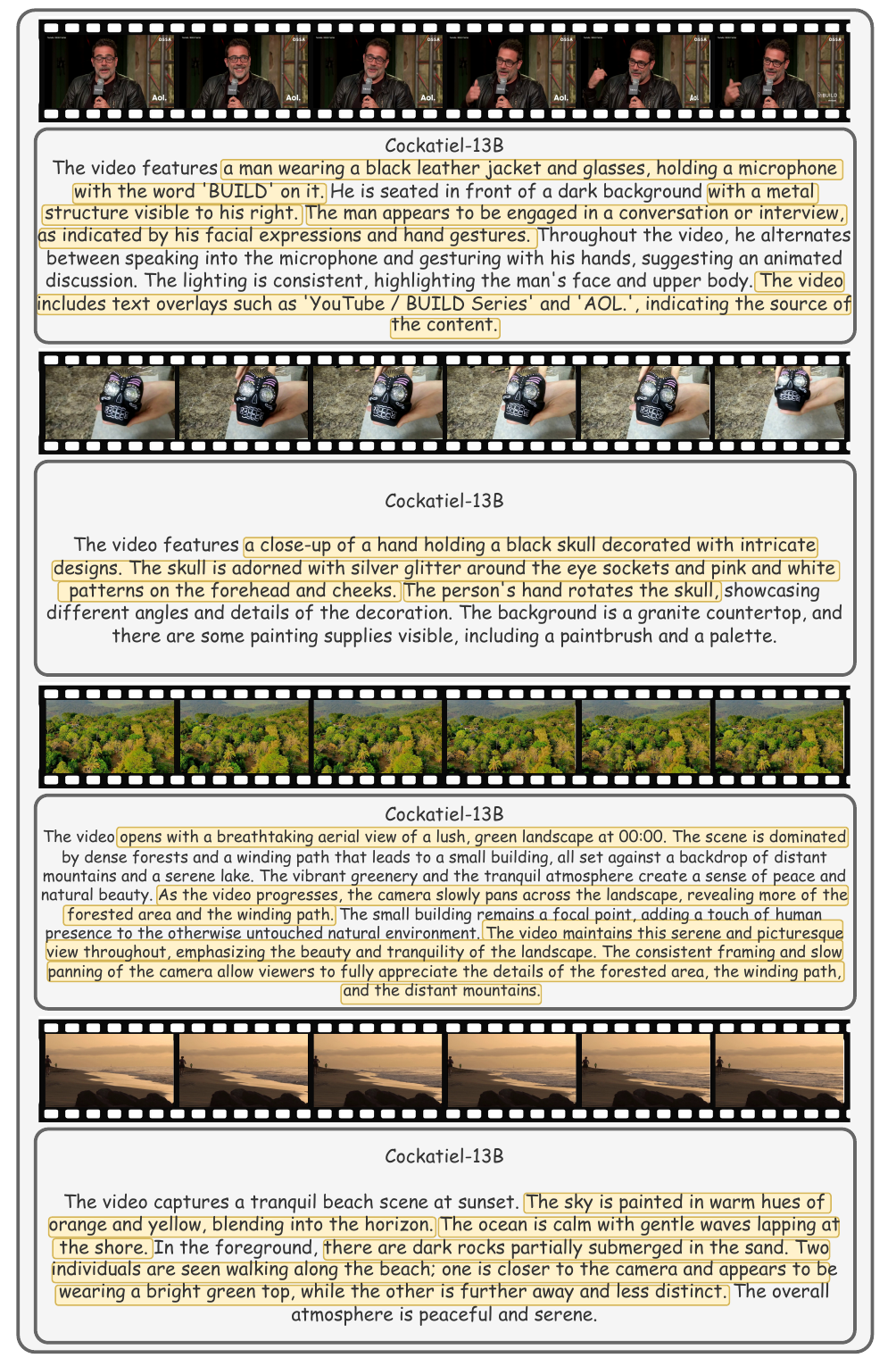}
\caption{
    More video detailed captions generated by Cockatiel-13B.
    The caption content that is captured by our models and align with the detailed visual elements in the videos are emphasized using yellow backgrounds.
}
\label{fig:supp_case_study}
\end{figure*}

\end{document}